\documentclass{article}

\usepackage[final,nonatbib]{tackling_climate_workshop_style}

\usepackage[utf8]{inputenc} 
\usepackage[T1]{fontenc}    
\usepackage{hyperref}       
\usepackage{url}            
\usepackage{booktabs}       
\usepackage{amsfonts}       
\usepackage{nicefrac}       
\usepackage{microtype}      
\usepackage{xcolor}         
\usepackage{graphicx}
\usepackage{subfigure}

\title{Predicting Critical Biogeochemistry of the Southern Ocean for Climate Monitoring}

\author{
  Ellen Park\\
  EAPS, MIT\\
  MC\&G, WHOI\\
  \And
  Jae Deok Kim\\
  EAPS, MIT\\
  Geology \& Geophysics, WHOI\\
  \AND
  Nadege Aoki\\
  EAPS, MIT\\
  Biology, WHOI\\
  \And
  Yumeng Melody Cao\\
  EECS\\
  MIT\\
  \And
  Yamin Arefeen\\
  EECS\\
  MIT\\
  \And
  Matthew Beveridge\\
  EECS\\
  MIT\\
  \And
  David Nicholson\\
  Marine Chemistry and Geochemistry\\
  WHOI\\
  \And
  Iddo Drori\\
  EECS\\
  MIT
}

\begin{document}

\maketitle

\begin{abstract}
  The Biogeochemical-Argo (BGC-Argo) program is building a network of globally distributed, sensor-equipped robotic profiling floats, improving our understanding of the climate system and how it is changing. These floats, however, are limited in the number of variables measured. In this study, we train neural networks to predict silicate and phosphate values in the Southern Ocean from temperature, pressure, salinity, oxygen, nitrate, and location and apply these models to earth system model (ESM) and BGC-Argo data to expand the utility of this ocean observation network. We trained our neural networks on observations from the Global Ocean Ship-Based Hydrographic Investigations Program (GO-SHIP) and use dropout regularization to provide uncertainty bounds around our predicted values. Our neural network significantly improves upon linear regression but shows variable levels of uncertainty across the ranges of predicted variables. We explore the generalization of our estimators to test data outside our training distribution from both ESM and BGC-Argo data. Our use of out-of-distribution test data to examine shifts in biogeochemical parameters and calculate uncertainty bounds around estimates advance the state-of-the-art in oceanographic data and climate monitoring. We make our data and code publicly available.
\end{abstract}

\section{Introduction}
\vspace{-8pt}
Ship-based ocean measurements, like those collected by the Global Ocean Ship-Based Hydrographic Investigations Program (GO-SHIP), provide valuable insight into ocean carbon uptake, biological processes, circulation, and climate variability. However, research cruises are expensive, sparse and often seasonally biased due to weather conditions. The Biogeochemical-Argo (BGC-Argo) program aims to become the first globally comprehensive sensing array for ocean ecosystems and biogeochemistry. Yet, profiling floats are limited in the number of sensors they can support \cite{chai2020monitoring}. Developing models which accurately predict additional features, such as nutrient ratios, from limited sensor data will broaden the applicability of BGC-Argo floats and allow us to better monitor and understand changes to the Earth's climate. 

Previous work demonstrates the utility of applying machine learning to cruise and float data to estimate values of global N$_2$ fixation \cite{tang2019machine}, particulate organic carbon \cite{sauzede2020estimation}, alkalinity, pH, and nitrate \cite{carter2018updated}. Bittig et al. \cite{bittig2018alternative} demonstrate advantages of using Bayesian neural networks, which account for uncertainty around predicted values, to estimate nutrient concentrations, and D’Alelio et al. \cite{d2020machine} show an application of regression methods for examining interannual variability in primary productivity. 

We draw on these methods to develop neural networks trained on cruise data to predict phosphate and silicate, important nutrients controlling ocean productivity and biodiversity \cite{weber2010ocean}. This is important because these nutrients regulate biological processes that remove carbon from the surface ocean at an annual rate roughly equivalent to annual anthropogenic carbon dioxide emissions \cite{siegel2014},\cite{NOAAwebsite}. The Southern Ocean is selected for developing and testing these models as it is an important global carbon sink and has the most extensive BGC-Argo float coverage at this time \cite{gruber2019variable}. 

\begin{figure}[t!]
\label{fig:goship}
    \centering
    \begin{subfigure}{}
        \includegraphics[height=2.0in]{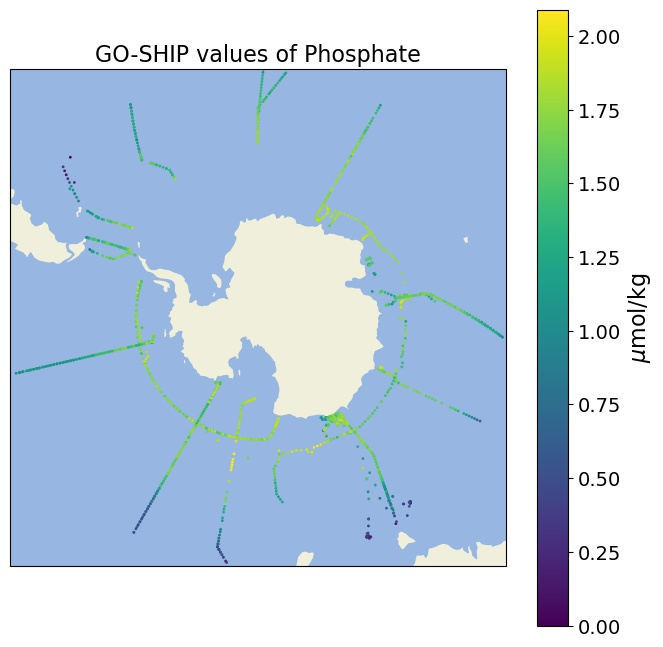}
    \end{subfigure}
    \begin{subfigure}{}
        \includegraphics[height=2.0in]{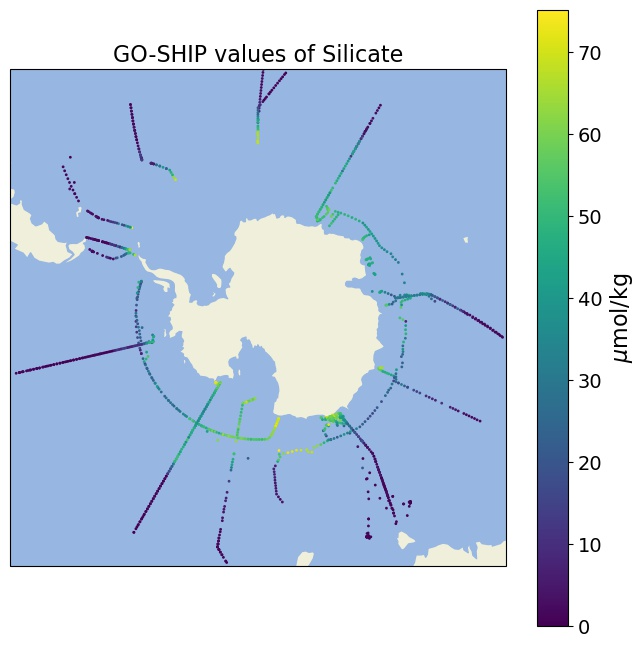}
    \end{subfigure}
    \caption{Transect locations of GO-SHIP oceanographic cruises in the Southern Ocean, between 03/08/2001-05/02/2019. Latitude 45-90º S, Longitude: -180-180º E, with surface (P < 10 dbar) values of phosphate (left) and silicate (right) in $\mu$mol kg$^{-1}$.}
\end{figure}

\section{Methods}
\vspace{-8pt}
\paragraph{Dataset and training.} We use GO-SHIP data \cite{goship-data} in our training set to train our models. The data set includes 42,412 data points from Southern Ocean cruises between 2001-2019 as shown in Figure \ref{fig:goship}. We use GO-SHIP data for latitude, longitude, pressure, temperature, salinity, oxygen and nitrate to predict phosphate and silicate. We restrict our data to latitudes below 45º S, remove rows with missing data and furthermore follow the standards of the World Ocean Circulation Experiment Hydrographic Program and use quality control (QC) flags to down-select our data. We standardize the pressure, temperature, salinity, oxygen, and nitrate features. The position latitude and longitude data is projected to the WGS 84 / Antarctic Polar Stereographic coordinate reference system. We do not include time dependency (month) because initial evaluation of our linear regression indicates low importance of seasonal variability in predicting silicate and phosphate variation. We randomly shuffle the feature encoded data into a 9:1 ratio of training to test size and train our model using 10-fold cross-validation with mean-squared error loss. We select the model with lowest validation loss to evaluate the testing error for both phosphate and silicate.
\vspace{-8pt}
\paragraph{Model.} To evaluate uncertainty when predicting silicate and phosphate from our data, we train a i) a 1-layer feed-forward, fully connected neural network with linear activation (equivalent computation to linear regression) and ii) 2-layer feed-forward, fully-connected neural network with 64 hidden units, ReLU activation, and $p = 0.2$ dropout probability. We estimate uncertainty by sampling using dropout \cite{kendall2017uncertainties}, training the network using dropout and then testing each example by running multiple forward passes with dropout weights.
\vspace{-8pt}
\paragraph{Model applications: ESM and BGC-Argo data.} We evaluate our networks performance by comparing our model's results of phosphate and silicate to the values predicted from an Earth system model (ESM). We use the Institut Pierre Simon Laplace Climate Model 5 (IPSL-CM5) \cite{ipsl-cm5-data} model results from a 10-year historical model run initialized in 2000 and a 30-year projection initialized in 2005. We take the monthly-averaged surface values (59,088) of temperature, salinity, oxygen, nitrate, phosphate, and silicate at each location over the historical and predicted span of 35 years (2000-2035), apply our network model to these surface values (assuming surface pressure = 5 dbar), and compare our model results to the IPSL-CM5 values of phosphate and silicate. 

Next, we apply our network to test data from BGC-Argo float profiles located in the Southern Ocean equipped with both dissolved oxygen and nitrate sensors. There are 175 floats between 2000-2020, measuring a total of 16,710 profiles that meet these criteria, and we only use data points where all input features are measured. We apply our network to 181,329 data points and run 100 dropout iterations to generate standard deviations for our estimates.

\section{Results}
\begin{figure}
    \centering
    \includegraphics[width = .55\linewidth]{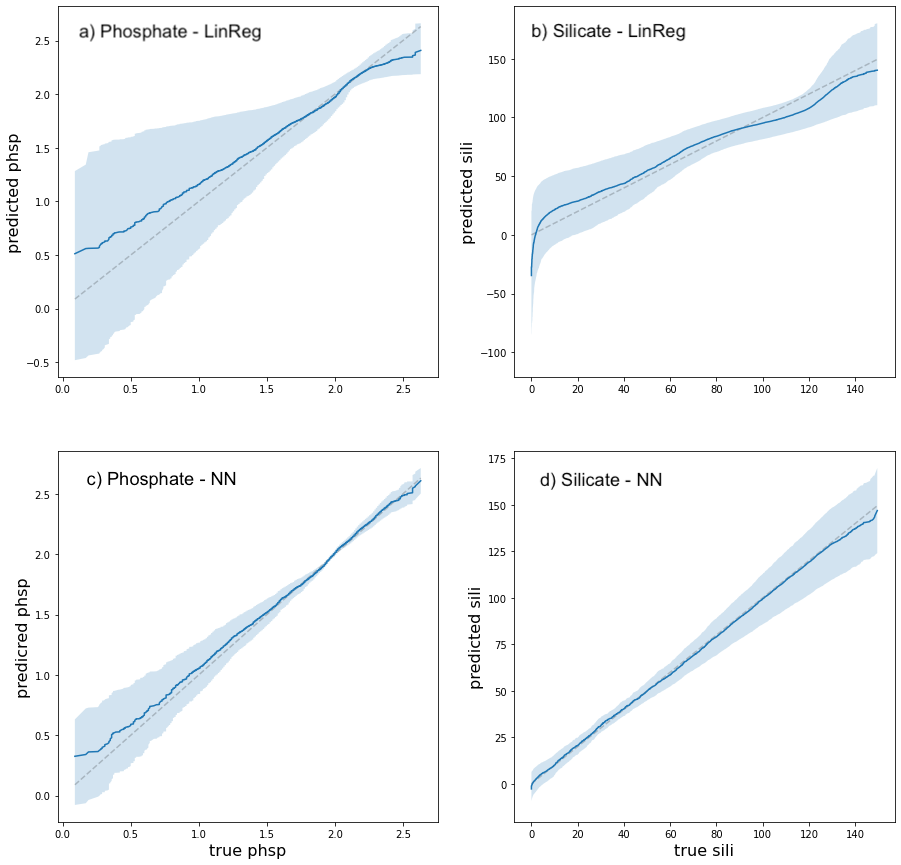}
    \caption{Results of uncertainty calculations using dropout regularization for linear regression based on OLS and a neural network, with plots for phosphate shown in (a) and (c) and plots for silicate shown in (b) and (d). Blue lines show the mean predicted values over 100 iterations and shaded areas represent 95\% confidence intervals. The dotted gray line denotes ground truth.}
    \label{fig:cf_subplots}
    \vspace{-10pt}
\end{figure}

\vspace{-8pt}
\paragraph{Comparing linear regression with neural networks}
The results from our linear regression analysis reveal a greater uncertainty in our estimated phosphate values compared to our silicate values. Additionally, the uncertainty of our silicate results are more uniform over our test data range, while the phosphate results have a greater uncertainty at lower values and lower uncertainty at higher ones (see Figure \ref{fig:cf_subplots} top row). The uncertainties in our phosphate and silicate estimates are reduced with our two-layer neural network (see Figure \ref{fig:cf_subplots} bottom row). The mean squared error also decreases substantially for both phosphate (MSE Linear: 0.019, MSE NN: 0.0031) and silicate (MSE Linear: 240, MSE NN: 50). The greatest uncertainties for phosphate are at lower values. For silicate, the greatest uncertainties are at higher silicate values. This could be a result of the differences in distribution of these compounds in the water column. Phosphate has greater variance in the upper water column (where phosphate values are lowest) and lower variance at depth, while the variance of silicate is more uniform throughout the water column. 

\vspace{-8pt}
\paragraph{Neural networks for ESM data.} We compared the ESM output values of phosphate and silicate to our neural-network predicted values of phosphate and silicate from the ESM features (see Figure 4 in Appendix). Our neural network under predicts phosphate values across the Southern Ocean and under predicts silicate values away from the Antarctic continent compared to the ESM values. However, our neural network is able to capture the spatial variations for both surface phosphate and silicate. These results suggest that our neural network model is able to capture processes modeled by the ESM. However, there are still discrepancies between these two model types. Based on these results, we believe our neural network has a high enough performance to apply to BGC-Argo data to estimate phosphate and silicate values from true observations. 

\begin{figure}[t!]
    \centering
    \label{fig:argo}
    \begin{subfigure}{}
        \includegraphics[height=3.6in]{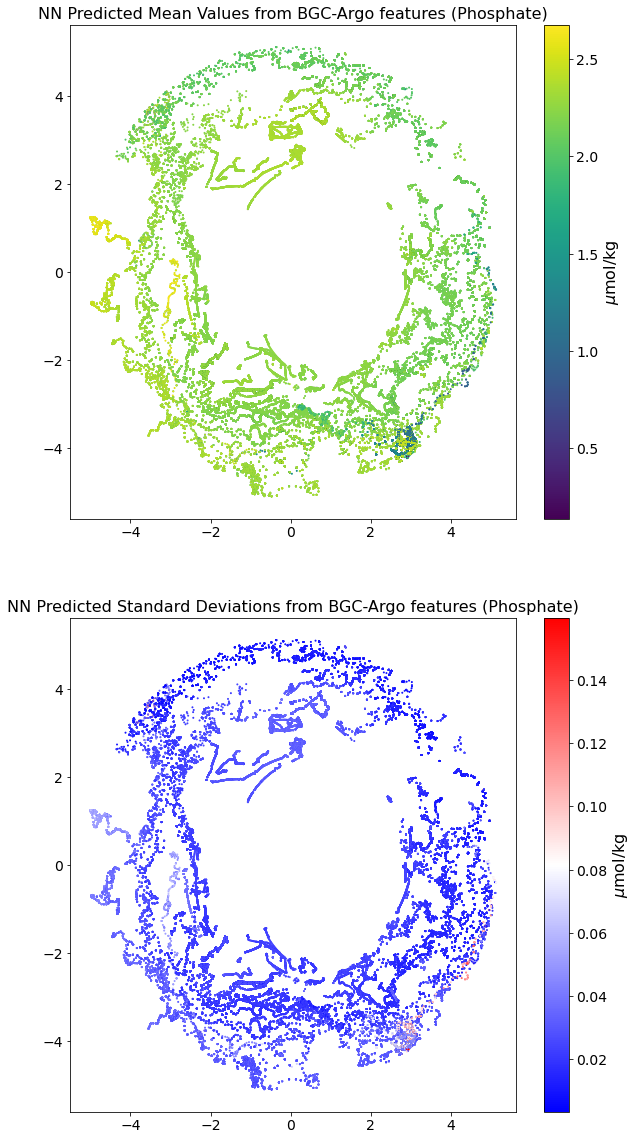}
    \end{subfigure}
    \begin{subfigure}{}
        \includegraphics[height=3.6in]{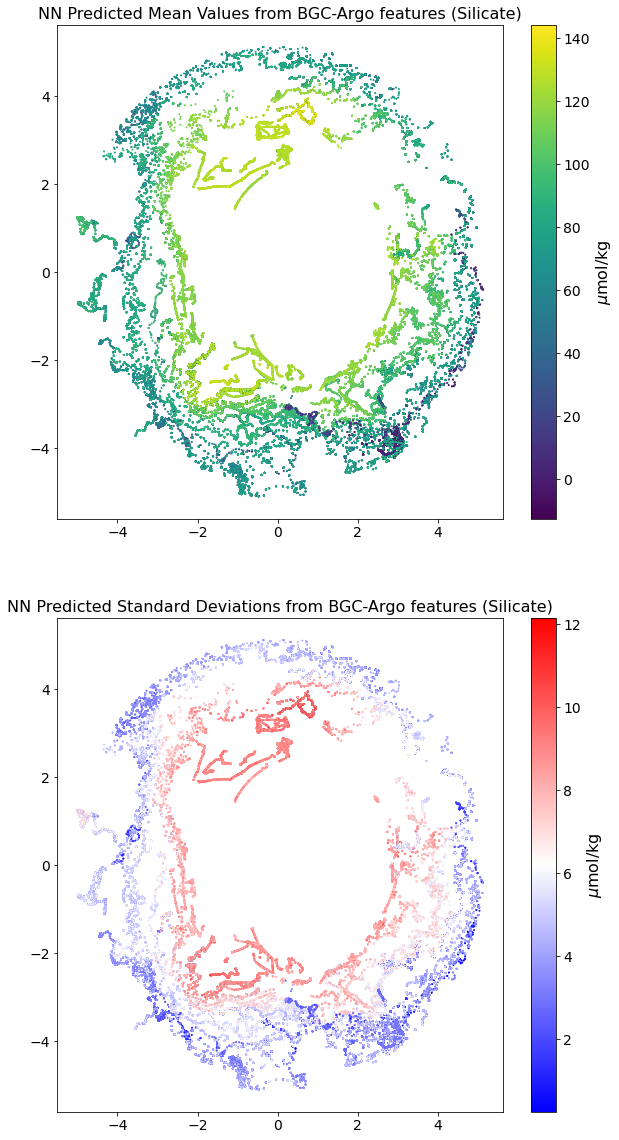} 
    \end{subfigure}
    \caption{Results from applying our neural network to predict phosphate and silicate ($\mu$mol kg$^{-1}$) using BGC-Argo float data. Top row: Predicted mean values for phosphate (left) and silicate (right). Middle row: Predicted standard deviations for phosphate (left) and silicate (right). There is systematically higher uncertainty near the continent for silicate values.}
    \vspace{-10pt}
\end{figure}

\vspace{-8pt}
\paragraph{Neural networks for BGC-Argo data.} Our neural network applied to BGC-Argo data predicts similar spatial patterns of phosphate and silicate to those measured by GO-SHIP and modeled by the ESM (see Figure \ref{fig:argo}). However, there are a few float trajectories that have noticeably different values from other floats in the region. While this could be due to local biogeochemical processes, it likely is due to sensor noise or drift that was missed during quality control. The uncertainties estimated for phosphate are generally low and uniform throughout the region, while the uncertainty estimates for silicate present similar spatial patterns as the mean value estimates, with high uncertainties near the continent. This suggests that there is a systematic error close to the continent, which could be attributed to ice dynamics causing higher variability in our features. These results suggest a relationship between latitude and silicate distributions. 

\vspace{-8pt}
\paragraph{Limitations.} Our neural network models are generally successful, demonstrating high potential for progress in this application. However, our proof-of-concept implementation leaves areas for improvement. We plan to improve our models by: (i) including a temporal component and using a spatial-temporal graph neural network representation; (ii) preserving the spatial relationships within the training data using a graph neural network; and (iii) training the models on a subset of shallower GO-SHIP data to better compare our model output to the surface model results from the ESM.

\vspace{-8pt}
\section{Conclusions}
\vspace{-8pt}
Our neural networks, trained on ship-based measurements using dropout regularization to predict dissolved phosphate and silicate values in the Southern Ocean, performed better than our linear regression models. However, the neural networks showed variable levels of uncertainty across the ranges of both predicted variables. When we apply our neural network models to the ESM data, we are able to evaluate our empirical models' performance relative to the bio-physical models in the surface ocean. We then further apply the neural networks to the BGC-Argo data to demonstrate the models' goal application and to generally compare the results to the GO-SHIP and ESM data. Ultimately, our models are successful and provide a workflow that may be used and scaled globally and applied to the expanding BGC-Argo fleet. Furthermore, our results demonstrate the utility of applying machine-learning based regression methods to estimate additional biogeochemical parameters from a limited set of oceanographic variables. This provides additional means for the scientific community to leverage existing earth system observational programs to monitor the climate system and its changes. All code is written in Python and shared online in an  {anonymous Dropbox}\footnote{\url{https://www.dropbox.com/sh/tpr3rw2i5g1fmdw/AABBZHyZ8erp_1hy6iEutix8a?dl=0}}. The ESM \cite{esm-data} and BGC-Argo data \cite{bgc-argo-data} are publicly available.

\section*{Acknowledgements}
We acknowledge model output from IPSL via CMIP and float data from the International Argo Program and the national programs that contribute to it. The Argo Program is part of the Global Ocean Observing System. Argo (2021). DN was supported by NSF$\#$1946578. EP was supported by NSF$\#$1756613. ID thanks Google for a cloud educational grant.

\bibliographystyle{plain}
\bibliography{bibliography}

\newpage
\clearpage

\section*{Appendix}
\begin{figure}[h!]
\label{fig:ipsl-cms}
    \begin{subfigure}{}
        \includegraphics[width = 0.5\textwidth]{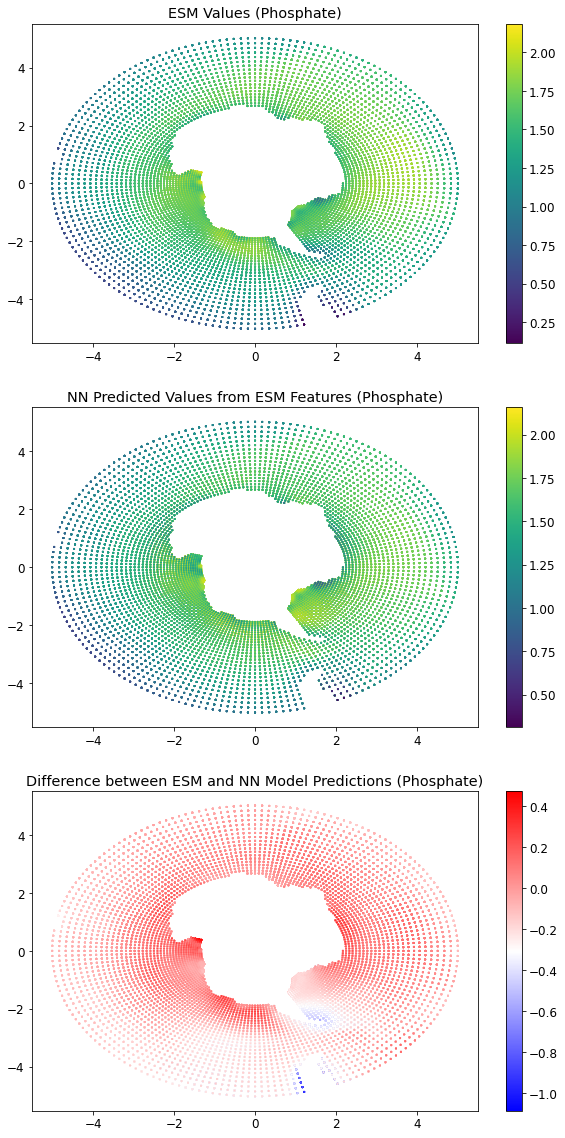}
    \end{subfigure}
    \begin{subfigure}{}
        \includegraphics[width=0.5\textwidth]{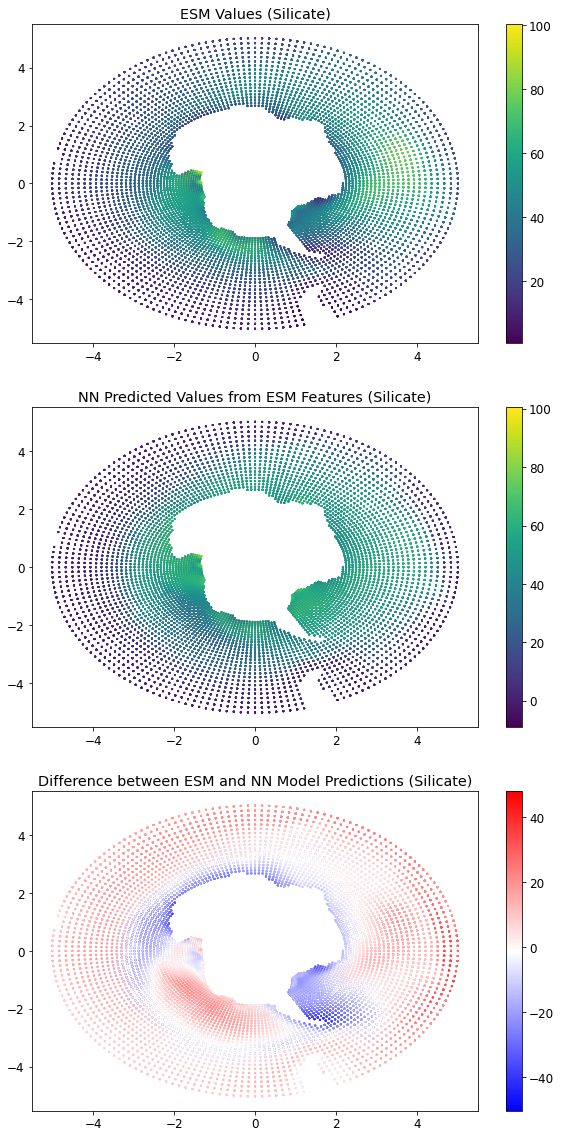}
    \end{subfigure}
    \caption{IPSL-CM5 model values for parameters of interest and results from applying our neural network to predict phosphate and silicate in this model space. Top row: Unchanged values from IPSL-CM5 model. Middle row: Predicted values from our neural network using ESM predictions for all features. Bottom row: ESM - NN. These figures show general agreement between these two approaches, with a systematic underestimation of phosphate by our neural network compared to the ESM and greater disagreement on both variables near the continent.}
\end{figure}

Figure \ref{fig:ipsl-cms} demonstrates our neural network model output values of phosphate and silicate to our neural-network predicted values of phosphate and silicate from the ESM features. Our neural network under predicts phosphate values across the Southern Ocean and under predicts silicate values away from the Antarctic continent compared to the ESM predicted values. However, our neural network is generally able to capture the spatial variations in both phosphate and silicate values. These results suggest that our neural network model is able to capture processes modeled by the ESM.

\end{document}